\documentclass[letterpaper, 10 pt, journal, twoside]{IEEEtran}
\usepackage{cite}
\usepackage{graphicx}
\usepackage{booktabs}
\usepackage{tabularx}
\usepackage{array}
\usepackage{color}
\usepackage{tabularray}
\UseTblrLibrary{booktabs}
\usepackage{xcolor}
\usepackage{gensymb}
\usepackage{threeparttable}
\usepackage{amsmath,amssymb,amsfonts,dsfont}
\usepackage{mathtools, bm}
\usepackage{algorithm,algorithmicx,listings}
\usepackage[noend]{algpseudocode}
\usepackage{hyperref}
\usepackage{float}
\usepackage{pifont}

\begin{document}

\thispagestyle{empty}
\begin{center}
	\vspace*{2em}
	\begin{minipage}{0.95\textwidth}
		{\small
			© 2025 IEEE. Personal use of this material is permitted.  
			Permission from IEEE must be obtained for all other uses, in any current or future media, including reprinting/republishing this material for advertising or promotional purposes, creating new collective works, for resale or redistribution to servers or lists, or reuse of any copyrighted component of this work in other works.
			
			\vspace{1em}
			
			This work has been accepted for publication in IEEE Robotics and Automation Letters (RA-L).  
			The final version is available at: \texttt{https://doi.org/10.1109/LRA.2025.3575320}
		}
	\end{minipage}
\end{center}

\newpage

\title{EKF-Based Radar-Inertial Odometry with Online Temporal Calibration}

\author{Changseung Kim$^{1}$, Geunsik Bae$^{1}$, Woojae Shin$^{1}$, Sen Wang$^{2}$, and Hyondong Oh$^{1}$%
\thanks{Manuscript received: January 25, 2025; Revised: April 14, 2025; Accepted: May 11, 2025.}
\thanks{This paper was recommended for publication by Editor Javier Civera upon evaluation of the Associate Editor and Reviewers’ comments.
This research was supported by the Technology Innovation Program (No. 20018110, "Development of a wireless teleoperable relief robot for detecting searching and responding in narrow space") funded By the Ministry of Trade, Industry \& Energy (MOTIE, Korea). \textit{(Corresponding author: Hyondong Oh)}} 
\thanks{$^{1}$Department of Mechanical Engineering, Ulsan National Institute of Science and Technology (UNIST), Ulsan 44919, Republic of Korea (e-mail: \{pon02124; baegs94; oj7987; h.oh\}@unist.ac.kr).}%
\thanks{$^{2}$Department of Electrical and Electronic Engineering, Imperial College London, SW7 2AZ London, United Kingdom (e-mail: sen.wang@imperial.ac.uk).}%
\thanks{Digital Object Identifier (DOI): see top of this page.}
}

\markboth{IEEE Robotics and Automation Letters. Preprint Version. Accepted May, 2025}
{Kim \MakeLowercase{\textit{et al.}}: EKF-Based Radar-Inertial Odometry with Online Temporal Calibration}

\maketitle
\begin{abstract}
Accurate time synchronization between heterogeneous sensors is crucial for ensuring robust state estimation in multi-sensor fusion systems. Sensor delays often cause discrepancies between the actual time when the event was captured and the time of sensor measurement, leading to temporal misalignment (time offset) between sensor measurement streams. In this paper, we propose an extended Kalman filter (EKF)-based radar-inertial odometry (RIO) framework that estimates the time offset online. The radar ego-velocity measurement model, derived from a single radar scan, is formulated to incorporate the time offset into the update. By leveraging temporal calibration, the proposed RIO enables accurate propagation and measurement updates based on a common time stream. Experiments on both simulated and real-world datasets demonstrate the accurate time offset estimation of the proposed method and its impact on RIO performance, validating the importance of sensor time synchronization. Our implementation of the EKF-RIO with online temporal calibration is available at \href{https://github.com/spearwin/EKF-RIO-TC}{https://github.com/spearwin/EKF-RIO-TC}.
\end{abstract}


\IEEEpeerreviewmaketitle

\section{INTRODUCTION}
\label{sec: introduction}
\IEEEPARstart{A}{ccurate} and robust state estimation is crucial for the successful execution of autonomous missions using mobile robots or vehicles. Global navigation satellite system provides reliable estimation in typical outdoor environments, but its reliability degrades or it is unavailable in obstructed areas such as urban canyons or indoor environments. In such cases, alternative state estimation methods such as simultaneous localization and mapping (SLAM) or odometry using exteroceptive sensors (e.g., camera, LiDAR, and radar) are crucial to maintain reliable autonomy.

Real-time state estimation using light-based sensors such as cameras and LiDAR has significantly improved over the past two decades~\cite{9196524, 9440682, 9341176, 9697912}. Vision-based methods show notable performance across a wide range of conditions, despite their reliance on small and lightweight sensors. However, their performance significantly decreases in environments with lighting changes or visually featureless surfaces~\cite{zhang2018laser}. In contrast, LiDAR-based methods are resilient to lighting conditions and can accurately capture the detailed structure of the surrounding environment over long distances. However, they struggle in self-similar or structureless environments, such as long corridors or flat planes~\cite{10611444}. Moreover, light-based sensors face limitations when exposed to small particles, such as snow, fog, or dust, due to their short wavelength.

\begin{figure}[t]
	\centering
	\includegraphics[width=\linewidth]{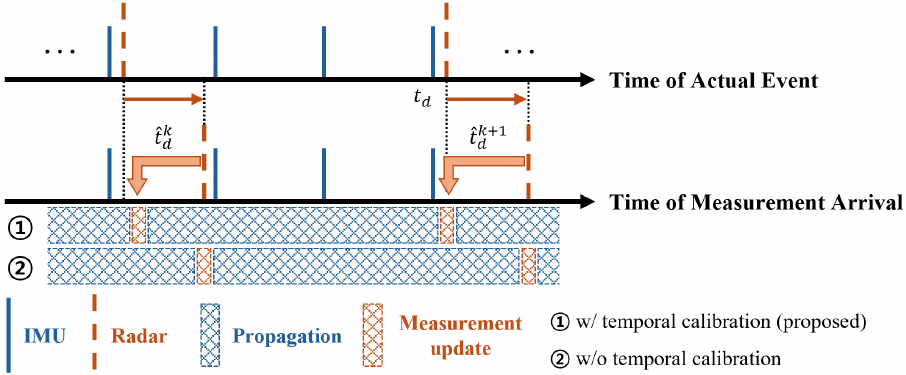}
	\vspace{-2.0em}
	\caption{Illustration of the temporal misalignment between IMU and radar measurement streams, along with the corresponding EKF propagation and measurement update steps. The time offset between sensors is denoted as $t_d=t_{d,IMU}-t_{d,Radar}$. For simplicity, the IMU delay is omitted in the figure. Two approaches are shown: with temporal calibration (proposed) and without it.}
	\vspace{-0.5em}
	\label{fig1}
\end{figure}

Recently, a radar has gained attention as a promising solution to address these challenges~\cite{10683889}. In particular, a frequency-modulated continuous wave (FMCW) millimeter-wave 4D radar not only penetrates small particles effectively due to its relatively long wavelength but also measures relative speed (i.e., doppler velocity) to surrounding environments through frequency modulation. This radar typically provides measurements at a rate of 5-20 Hz, including raw signal data, 3D point clouds with spatial information and Doppler velocity for each point, enabling the estimation of ego-velocity from a single radar scan~\cite{6728341}. However, due to its relatively low sensor rate, it is necessary to predict the movement between consecutive radar scans. To address this, sensor fusion with an IMU, which operates at a higher sensor rate of over 100 Hz, can be utilized. The IMU complements the movement between consecutive radar scans, improving overall estimation accuracy and robustness. Therefore, there has been increasing interest and active research in radar-inertial odometry (RIO), including both loosely and tightly coupled methods, which integrate the IMU with the radar.

Accurate time synchronization between heterogeneous sensors is crucial for data fusion. All sensors inherently experience delays, leading to a discrepancy between the actual time when the event was captured and the time when the measurement is timestamped, as illustrated in Fig.~\ref{fig1}. While IMU systems typically have minimal latency, radar systems experience more significant delays due to factors such as signal processing steps, including fast Fourier transform, beamforming techniques, detection algorithms, and elevation angle estimation. Additionally, inherent hardware delays, including the analog-to-digital converter start-time and transmission delay, further contribute to the overall latency of radar systems~\cite{10477463}. Differences in sensor delays lead to temporal misalignments, which can pose significant challenges in multi-sensor systems. To address this issue, special hardware triggers or manufacturer software support may be required, but not all sensors provide such functions~\cite{10610666}. The time offset, which represents the difference in delays between IMU and radar systems, can reach up to hundreds of milliseconds, posing significant challenges for accurate data fusion in RIO. Compared with LiDAR-IMU and camera-IMU systems, our experimental results show that radar-IMU systems have a significantly larger time offset, highlighting the importance of temporal calibration.

In this paper, we propose a method for real-time estimation of the time offset between the IMU and the radar in an extended Kalman filter (EKF)-based RIO. Unlike existing RIO studies that rely on hardware/software triggers or do not consider the time offset, our method directly estimates the time offset online by utilizing the radar ego-velocity. The proposed method demonstrates efficient and robust performance across multiple datasets. Additionally, we show that accounting for the time offset in RIO enhances overall accuracy, even when using the same measurement model. The main contributions of this study can be summarized as: 
\begin{enumerate}
    \item We propose an EKF-RIO-TC framework that estimates the time offset between the IMU and the radar in real-time, utilizing the radar ego-velocity estimated from a single radar scan;
    \item The proposed method is validated through both simulated and real-world datasets with and without hardware triggers. The results show that the time offset between sensors is non-negligible and must be accurately estimated to improve RIO performance; and
    \item To benefit the community, the implementation code has been made open-source. The proposed method is easy to implement and applicable, as it utilizes the radar ego-velocity measurement, which is commonly used in most RIO studies. To the best of our knowledge, this is the first work to implement online temporal calibration between sensors in RIO.
\end{enumerate}

The remainder of this paper is organized as follows. Section~\ref{sec: related work} reviews the related work relevant to our research. Section~\ref{sec: notation} outlines notations used throughout the paper. Section~\ref{sec: filter description} details the framework of the proposed RIO, accounting for the time offset between sensors. Section~\ref{sec: experiments} validates the proposed method across multiple datasets and analyzes the results. Finally, Section~\ref{sec: conclusion} summarizes the findings and concludes the paper.
\section{Related Work}
\label{sec: related work}
In this section, we introduce filter-based RIO research using FMCW 4D radar. Because no existing research considers temporal calibration in the radar-IMU systems, we subsequently introduce research into online temporal calibration in other multi-sensor fusion systems.

\subsection{Radar-Inertial Odometry with FMCW 4D Radars}
Doer and Trommer \cite{9235254} proposed an EKF-based RIO framework that fuses IMU data with 3D ego-velocity estimated from radar measurements. They introduced the 3-point RANSAC-LSQ to remove outliers in ego-velocity estimation. Later, the same authors extended the filter state to include the extrinsic parameters between the IMU and the radar for online extrinsic estimation \cite{9317343}, and enhanced yaw estimation for indoor environments using Manhattan world assumptions \cite{9470842}. Michalczyk et al. \cite{9981396} focused on the relatively accurate radar measurements, specifically distance and Doppler velocity, utilizing them for filter updates. In addition to ego-velocity updates, they constructed point-to-point residuals using distance as the most informative dimension for updates. The same authors extended this approach by incorporating multi-pose into the framework and including long-term observed points in the filter state to further improve performance \cite{10160482}. Zhuang et al. \cite{10100861} adopted graduated non-convexity for ego-velocity estimation and applied distribution-to-multi-distribution constraints for sparse scan matching. They also incorporated scancontext for place revisits and loop closures via pose graphs. Kubelka et al. \cite{10610666} compared scan matching-based methods (e.g., ICP, APD-GICP, and NDT) and radar ego-velocity-based methods (e.g., IMU attitude with radar ego-velocity and EKF-RIO \cite{9235254}) across two distinct radar sensor setups. The former showed better in dense point cloud setups, while the latter outperformed in sparse setups. In the sparse radar setups, scan matching often failed due to low point cloud density. Performance degradation in radar ego-velocity-based methods was also observed in specific sections, such as when the vehicle hit a bump, causing discrepancies between the IMU and the radar measurements due to the temporal misalignment. Their analysis on this issue motivated our research.

In \cite{9235254, 9317343, 9470842}, hardware triggers were implemented on a microcontroller board for sensor time synchronization. In \cite{10610666}, and~\cite{9981396, 10160482, 10100861}, the authors did not account for the time offset when constructing their system frameworks, assuming the times of sensor measurements were accurate. Our proposed method estimates the time offset in real-time from radar ego-velocity, without relying on physical triggers. Since the ego-velocity is estimated from a single scan, there is no need for matching between consecutive scans, making the method independent of radar point cloud density and thereby offering greater flexibility for various radar sensors. Furthermore, since all the mentioned works~\cite{10610666, 9317343, 9235254, 9470842, 9981396, 10160482, 10100861} utilize radar ego-velocity in the measurement update, the proposed method can be seamlessly integrated into their frameworks, not only ensuring accurate time synchronization but also potentially improving scan matching accuracy.

\subsection{Online Temporal Calibration for Multi-Sensor Fusion Systems}
Qin and Shen \cite{8593603} proposed an optimization-based method for visual-inertial odometry (VIO) that enables online temporal calibration. They addressed time synchronization by jointly optimizing prior, IMU, and vision factors which account for the time offset between sensors. In particular, they compensated for feature measurements using their velocities on the image plane along with the time offset. Li and Mourikis \cite{li2014online} proposed a filter-based method to estimate the time offset between a camera and an IMU by including the time offset variable in the filter state. They formulated the camera measurement models for the 3D feature positions using the filter state including the time offset, thereby effectively estimating the time offset and improving the performance of VIO. Lee et al. \cite{9561254} proposed a filter-based method for multi-sensor fusion odometry involving a LiDAR and an IMU. Since finding point correspondences between LiDAR scans can be challenging, they used plane patches to handle the data more efficiently. Their method involved extracting plane patches from the point clouds and incorporating the time offset variable into the plane measurement models.

In contrast to previous works \cite{8593603, li2014online, 9561254} which estimate the time offset by matching features between consecutive images or scans, our proposed method estimates the time offset from a single radar scan by formulating the radar ego-velocity measurement models. Unlike the camera and the LiDAR, the radar uniquely measures the Doppler velocity, enabling the direct estimation of ego-velocity. Instead of relying on scan matching, which can be challenging with sparse radar data, the proposed method avoids potential inaccuracies associated with correspondence matching.
\section{Notation}
\label{sec: notation}
Uppercase letters in superscripts (e.g., $A$ in ${}^A\mathbf{q}_B$) denote the reference frame. Quaternions follow the Hamilton convention \cite{sola2017quaternion}. Vectors are represented by bold lowercase letters, matrices by bold uppercase letters, and scalars by non-bold lowercase letters.

${}^A\mathbf{q}_B$ represents the quaternion describing the attitude of frame $B$ relative to $A$. The corresponding rotation matrix ${}^A\mathbf{R}_B = \mathbf{R}({}^A\mathbf{q}_B)$ belongs to $\text{SO}(3)$. ${}^A\mathbf{p}_B$ represents the position of frame $B$ relative to $A$, expressed in $A$.
\section{Filter Description}
\label{sec: filter description}
The system uses three coordinate frames: global $G$, IMU $I$, and radar $R$. The proposed EKF-based RIO estimates the 6D pose of frame $I$ with respect to the global reference frame $G$. The estimator utilizes the error state extended Kalman filter (ES-EKF), which effectively handles nonlinear dynamics and measurement models in pose estimation. By maintaining a minimal error-state near the origin, the filter avoids over-parameterization and singularities, and simplifies Jacobian computations, enhancing consistency and efficiency when fusing IMU and radar measurements.

\subsection{System Overview}
\label{sec: system overview}
Figure~\ref{fig1} illustrates the temporal misalignment between the IMU and radar measurement streams, and the corresponding EKF propagation and update steps. The upper plot shows the actual time when the event was captured by the sensors, while the lower plot represents the time of measurement arrival, which corresponds to the timestamp assigned to the sensor data. Due to individual sensor delays (i.e., $t_{d,IMU}$ and $t_{d,Radar}$), the timestamp assigned to the measurement does not match the actual event time. The time offset $t_d$ represents the difference between the sensor delays, defined as:
\begin{equation}
\label{time_offset}
    t_d = t_{d,IMU} - t_{d,Radar}.
\end{equation}
Since the radar typically has a larger delay than the IMU, $t_d$ generally takes a negative value.

Traditional EKF-based RIO, which does not account for temporal calibration, performs propagation with IMU data and updates using radar measurements based on the time of measurement arrival. To ensure accurate state estimation, it is crucial to align the sensor measurements from both the IMU and the radar to a common time stream. While the exact delays of individual sensors are difficult to determine, the time offset can be estimated. The proposed method estimates the time offset $t_d$ in real-time using radar ego-velocity, and adjusts the radar measurement to align with the IMU measurement time stream, which serves as the common time reference. By leveraging temporal calibration, the proposed RIO enables propagation and measurement updates to be performed based on a common time stream.
\subsection{System State}
\label{sec: system state}
At time step \( k \), the system state is defined as:
\begin{equation}
    \mathbf{x}^k = 
    \left(
    {}^G\mathbf{q}_I^{k\top} \quad 
    \mathbf{b}_g^{k\top} \quad 
    {}^G\mathbf{v}_I^{k\top} \quad 
    \mathbf{b}_a^{k\top} \quad
    {}^G\mathbf{p}_I^{k\top} \quad
    {t}_d^k
    \right)^\top,
\end{equation}
where ${}^G\mathbf{q}_I$ represents the attitude, ${}^G\mathbf{v}_I$ the velocity, and ${}^G\mathbf{p}_I$ the position of the IMU. The terms $\mathbf{b}_g$ and $\mathbf{b}_a$ represent the gyroscope and accelerometer biases, respectively, and ${t}_d$ represents the time offset defined in Eq.~\eqref{time_offset}.

The error state formulation, as highlighted in \cite{sola2017quaternion}, minimizes errors and avoids parameter singularities. Given the estimated state $\hat{\mathbf{x}}$ and the error state $\tilde{\mathbf{x}}$, the true state $\mathbf{x}$ is expressed as:
\begin{equation}
    \mathbf{x} = \hat{\mathbf{x}} + \tilde{\mathbf{x}}.
\end{equation}
The true quaternion $\mathbf{q}$ is represented as a combination of the estimated quaternion $\hat{\mathbf{q}}$ and the error quaternion $\tilde{\mathbf{q}}$ as $\mathbf{q} = \hat{\mathbf{q}} \otimes \tilde{\mathbf{q}}$, where $\otimes$ denotes quaternion multiplication. The error quaternion $\tilde{\mathbf{q}}$ is approximated by $\tilde{\mathbf{q}} \approx \begin{bmatrix} 1 & \frac{1}{2} \boldsymbol{\theta}^\top \end{bmatrix}^\top$, with $\boldsymbol{\theta}$ representing a small Euler angle error.

Then, the error state at time step $k$ is similarly defined as:
\begin{equation}
    \tilde{\mathbf{x}}^k = 
    \left(
    {}^G\bm{\theta}_I^{k\top} \quad 
    \tilde{\mathbf{b}}_g^{k\top} \quad 
    {}^G\tilde{\mathbf{v}}_I^{k\top} \quad 
    \tilde{\mathbf{b}}_a^{k\top} \quad
    {}^G\tilde{\mathbf{p}}_I^{k\top} \quad
    \tilde{t}_d^k
    \right)^\top.
\end{equation}
For simplicity, the time index \(k\) is omitted in the following equations.
\subsection{Propagation with IMU Measurements}
\label{sec: Propagation}

The continuous-time dynamics for the estimated state is expressed as follows:
\begin{equation}
\label{propagation}
    \begin{gathered}
        {}^G\dot{\hat{\mathbf{q}}}_I = \frac{1}{2} \mathbf{\Omega}({}^I\hat{\boldsymbol{\omega}}_I){}^G\hat{\mathbf{q}}_I, \quad
        \dot{\hat{\mathbf{b}}}_g = \mathbf{0}_{3 \times 1}, \\
        {}^G\dot{\hat{\mathbf{v}}}_I = {}^G\hat{\mathbf{R}}_I{}^I\hat{\mathbf{a}}_I + {}^G\mathbf{g}, \quad
        \dot{\hat{\mathbf{b}}}_a = \mathbf{0}_{3 \times 1}, \\
        {}^G\dot{\hat{\mathbf{p}}}_I = {}^G\hat{\mathbf{v}}_I, \quad
        \dot{\hat{t}}_d = 0,
    \end{gathered}
\end{equation}
where ${}^G\mathbf{g}$ represents the gravity vector. The estimated angular velocity ${}^I\hat{\boldsymbol{\omega}}_I$ and acceleration ${}^I\hat{\mathbf{a}}_I$ are expressed as ${}^I{\hat{\boldsymbol{\omega}}}_I = {}^I{\boldsymbol{\omega}}_I^m - \hat{\mathbf{b}}_g$ and ${}^I\hat{\mathbf{a}}_I = {}^I{\mathbf{a}}_I^m - \hat{\mathbf{b}}_a$, where ${}^I{\boldsymbol{\omega}}_I^m$ and ${}^I{\mathbf{a}}_I^m$ denote the gyroscope and accelerometer measurements, respectively. The matrix $\mathbf{\Omega}(\hat{\boldsymbol{\omega}})$, constructed from the estimated angular velocity $\hat{\boldsymbol{\omega}}$ and its skew-symmetric matrix $\lfloor \hat{\boldsymbol{\omega}} \times \rfloor$, is represented as: 
\begin{equation}
    \mathbf{\Omega}(\hat{\boldsymbol{\omega}}) = 
    \begin{bmatrix}
        0 & -\hat{\boldsymbol{\omega}}^\top \\
        \hat{\boldsymbol{\omega}} & -\lfloor \hat{\boldsymbol{\omega}} \times \rfloor
    \end{bmatrix}.
\end{equation}
The estimated state $\hat{\mathbf{x}}$ is propagated with IMU measurements through the continuous-time dynamics in Eq.~\eqref{propagation}, using 4\textsuperscript{th}-order Runge-Kutta numerical integration.

For the covariance propagation, the linearized continuous-time dynamics for the error state is expressed as:
\begin{equation}
\label{ekf}
    \dot{\tilde{\mathbf{x}}} = \mathbf{F}\tilde{\mathbf{x}} + \mathbf{G}\mathbf{n},
\end{equation}
where $\mathbf{n} = \left( \mathbf{n}_g^\top, \mathbf{n}_{wg}^\top, \mathbf{n}_a^\top, \mathbf{n}_{wa}^\top, n_d \right)^\top$. The vectors $\mathbf{n}_g$ and $\mathbf{n}_a$ represent the Gaussian noise on the gyroscope and accelerometer, while $\mathbf{n}_{wg}$ and $\mathbf{n}_{wa}$ denote their bias random walks. The scalar $n_d$ models Gaussian noise (i.e., uncertainty) in the time offset.

The matrix $\mathbf{F}$ represents the linearized system dynamics, and $\mathbf{G}$ models the influence of the process noise on the error state. Only the non-zero elements of the matrix $\mathbf{F}$ are given as follows:
\begin{equation}
    \begin{aligned}
        \mathbf{F}(0:2, 0:2) &= -\lfloor ({}^I{\boldsymbol{\omega}}_I^m - \hat{\mathbf{b}}_g) \times \rfloor, \\
        \mathbf{F}(0:2, 3:5) &= -\mathbf{I}_{3}, \\
        \mathbf{F}(6:8, 0:2) &= -{}^G\hat{\mathbf{R}}_I \lfloor ({}^I{\mathbf{a}}_I^m - \hat{\mathbf{b}}_a) \times \rfloor, \\
        \mathbf{F}(6:8, 9:11) &= -{}^G\hat{\mathbf{R}}_I, \\
        \mathbf{F}(12:14, 6:8) &= \mathbf{I}_{3}.
    \end{aligned}
\end{equation}
Similarly, the non-zero elements of the matrix $\mathbf{G}$ are given as follows:
\begin{equation}
    \begin{aligned}
        \mathbf{G}(0:2, 0:2) &= -\mathbf{I}_{3}, \\
        \mathbf{G}(3:5, 3:5) &= \mathbf{I}_{3}, \\
        \mathbf{G}(6:8, 6:8) &= -{}^G\hat{\mathbf{R}}_I, \\
        \mathbf{G}(9:11, 9:11) &= \mathbf{I}_{3}, \\
        \mathbf{G}(15, 12) &= 1.
    \end{aligned}
\end{equation}

To propagate the covariance, the discrete-time transition matrix $\mathbf{\Phi}_k$ and process noise covariance matrix $\mathbf{Q}_k$, derived from Eq.~\eqref{ekf}, are defined as follows:
\begin{equation}
    \begin{gathered}
        \mathbf{\Phi}_k = \mathbf{\Phi}(t_{k+1}, t_k) = \exp{\left( \int_{t_k}^{t_{k+1}} \mathbf{F}(\tau) d\tau \right)}, \\
        \mathbf{Q}_k = \int_{t_k}^{t_{k+1}} \mathbf{\Phi}(t_{k+1}, \tau) \mathbf{G}\mathbf{Q}\mathbf{G}^\top \mathbf{\Phi}(t_{k+1}, \tau)^\top d\tau,
    \end{gathered}
\end{equation}
where $\mathbf{Q}$ is the continuous-time process noise covariance matrix. The propagated covariance matrix is expressed as:
\begin{equation}
    \mathbf{P}_{k+1|k} = \mathbf{\Phi}_k \mathbf{P}_{k|k} \mathbf{\Phi}_k^\top + \mathbf{Q}_k.
\end{equation}
Considering the time offset \( t_d \), propagation is repeated up to just before the time of the radar measurement aligned to the IMU measurement time stream.
\subsection{Measurement Update with Radar Measurements}
\label{sec: measurement update}
The FMCW 4D radar provides 3D point clouds, where each point includes a 3D position and a scalar Doppler velocity. The Doppler velocity represents the radial velocity of a target point, expressed as:
\begin{equation}
    v_d^i = -{}^R\mathbf{v}_R \cdot \frac{{}^R\mathbf{p}_f^i}{||{}^R\mathbf{p}_f^i||}, \quad \text{for } i = 1, \dots, N,
\end{equation}
where \( N \) is the total number of detected points, \( v_d^i \) denotes the Doppler velocity of the \( i \)-th point, \( {}^R\mathbf{p}_f^i \) is the position vector of the \( i \)-th point in the radar frame, and ${}^R\mathbf{v}_R$ denotes the radar ego-velocity. To estimate the radar ego-velocity \( {}^R\mathbf{v}_R \) from noisy measurements, various methods, such as RANSAC and m-estimator-based optimization, have been proposed. In this work, we adopt the 3-point RANSAC-LSQ \cite{9235254}, a simple yet robust method that efficiently eliminates outliers and estimates the radar ego-velocity \( {}^R\mathbf{v}_R \), which is used in the measurement update.

When the IMU and radar are rigidly connected, the radar ego-velocity measurement model can be expressed using the system state. As derived in \cite{9235254}, the radar ego-velocity is expressed as:
\begin{equation}
\label{ego_vel}
    \begin{aligned}
        {}^R\mathbf{v}_R(t) =& {}^R\mathbf{R}_I \left( {}^G\mathbf{R}_I^\top(t) {}^G\mathbf{v}_I(t) \right. \\
        &+ \left. \lfloor ({}^I\boldsymbol{\omega}_I^m(t) - \mathbf{b}_g(t)) \times \rfloor {}^I\mathbf{p}_R \right),
    \end{aligned}
\end{equation}
where the extrinsic parameters, \( {}^R\mathbf{R}_I \) and \( {}^I\mathbf{p}_R \), between the IMU and the radar are assumed to be pre-calibrated and constant.

To incorporate temporal calibration into the radar update, the proposed method evaluates the predicted radar ego-velocity at the IMU time stream, offset by the estimated time offset \( \hat{t}_d \). Accordingly, the residual \( \mathbf{r} \) is computed as the difference between the radar ego-velocity \( {}^R\mathbf{v}_R(t) \), estimated from the current radar measurements, and the predicted radar ego-velocity \( {}^R\hat{\mathbf{v}}_R(t') \). The residual is expressed as:
\begin{equation}
\label{residual}
    \mathbf{r} = {}^R\mathbf{v}_R(t) - {}^R\hat{\mathbf{v}}_R(t') = \mathbf{h}(\tilde{\mathbf{x}}) + \mathbf{n}_r.
\end{equation}
Here, \( t \) denotes the radar measurement timestamp, and \( t' = t + \hat{t}_d \) represents the corresponding IMU time, which serves as the reference time stream in the filter. The term \( \mathbf{n}_r \) denotes the noise of the measurement. The function \( \mathbf{h}(\tilde{\mathbf{x}}) \) is a nonlinear function that relates the state error \( \tilde{\mathbf{x}} \) to the radar ego-velocity measurement residual. For use in the EKF, this function is linearized with respect to the system state. The measurement Jacobian matrix \( \mathbf{H} \) is expressed as follows:
\begin{equation}
\begin{aligned}
    \mathbf{H} &= 
    \begin{bmatrix}
        \mathbf{H}_q & \mathbf{H}_{b_g} & \mathbf{H}_v & \mathbf{0}_{3 \times 3} & \mathbf{0}_{3 \times 3} & \mathbf{H}_{t_d}
    \end{bmatrix}, \\
    \mathbf{H}_q &= {}^R\mathbf{R}_I \lfloor {}^G\hat{\mathbf{R}}_I^\top {}^G\hat{\mathbf{v}}_I \times \rfloor,\\
    \mathbf{H}_{b_g} &= {}^R\mathbf{R}_I \lfloor {}^I\mathbf{p}_R \times \rfloor,\\
    \mathbf{H}_v &= {}^R\mathbf{R}_I {}^G\hat{\mathbf{R}}_I^\top.
\end{aligned}
\end{equation}
In Eq.~\eqref{ego_vel}, the time-varying states are \( {}^G\mathbf{R}_I \) and \( {}^G\mathbf{v}_I \). By applying the chain rule, the Jacobian \( \mathbf{H}_{t_d} \) is expressed as:
\begin{equation}
\begin{aligned}
\label{td_jacobian}
\mathbf{H}_{t_d} =& \frac{\partial \mathbf{h}\left(\tilde{\mathbf{x}}\left(\tilde{t'}\right)\right)}{\partial {}^G\boldsymbol{\theta}_I\left(\tilde{t'}\right)} 
\cdot \frac{\partial {}^G\boldsymbol{\theta}_I\left(\tilde{t'}\right)}{\partial \tilde{t'}} 
\cdot \frac{\partial \tilde{t'}}{\partial \tilde{t}_d} \\
&+ \frac{\partial \mathbf{h}\left(\tilde{\mathbf{x}}\left(\tilde{t'}\right)\right)}{\partial {}^G\tilde{\mathbf{v}}_I\left(\tilde{t'}\right)} 
\cdot \frac{\partial {}^G\tilde{\mathbf{v}}_I\left(\tilde{t'}\right)}{\partial \tilde{t'}} 
\cdot \frac{\partial \tilde{t'}}{\partial \tilde{t}_d} \\
=& \mathbf{H}_q \left( {}^I\boldsymbol{\omega}_I^m\left(t'\right) - \hat{\mathbf{b}}_g\left(t'\right) \right) \\
&+ \mathbf{H}_v\left( {}^G\hat{\mathbf{R}}_I\left(t'\right) \left({}^I\mathbf{a}_I^m\left(t'\right) - \hat{\mathbf{b}}_a\left(t'\right) \right) + {}^G\mathbf{g} \right).
\end{aligned}
\end{equation}

The EKF update proceeds by computing the Kalman gain \( \mathbf{K} \) as:
\begin{equation}
    \mathbf{K} = \mathbf{P}_{k+1|k} \mathbf{H}^\top \left( \mathbf{H} \mathbf{P}_{k+1|k} \mathbf{H}^\top + \mathbf{R} \right)^{-1},
\end{equation}
where \( \mathbf{R} \) represents the measurement noise covariance matrix. Finally, the estimated state and covariance are updated according to the Kalman gain as follows:
\begin{equation}
\begin{aligned}
    \hat{\mathbf{x}}_{k+1|k+1} &= \hat{\mathbf{x}}_{k+1|k} + \mathbf{K} \mathbf{r}, \\
    \mathbf{P}_{k+1|k+1} &= \left( \mathbf{I} - \mathbf{K} \mathbf{H} \right) \mathbf{P}_{k+1|k}.
\end{aligned}
\end{equation}
Each time a new radar measurement is received, the measurement update is performed based on the IMU time stream.
\subsection{Online Temporal Calibration}
\label{sec: online temporal calibration}
The proposed method estimates the time offset between the IMU and the radar in real-time by employing the radar ego-velocity measurement model. By accounting for the time offset, the proposed method ensures that both propagation and measurement updates are performed based on a common time stream, thereby synchronizing the measurements from both sensors.

The time offset is propagated using a noise model \( n_d \), as described in Eq.~\eqref{ekf}. If the time offset is constant over time or approximately known, it can be estimated without a noise model. However, the time offset varies across sensor models, making it difficult to predefine in most cases. For this reason, the time offset is modeled as a random walk.

In the measurement model presented in Eq.~\eqref{ego_vel}, the factors affected by the temporal misalignment between sensors are not only the state variables \( {}^G\mathbf{R}_I \), \( {}^G\mathbf{v}_I \), and \( \mathbf{b}_g \), but also the gyroscope measurement \( {}^I\boldsymbol{\omega}_I^m \). Although \( \mathbf{b}_g \), which does not change significantly over time, is negligible, failing to account for the time offset causes the state to be propagated using accelerometer and gyroscope measurements taken at incorrectly shifted times, offset by \( t_d \), as illustrated in Fig.~\ref{fig1}. This temporal misalignment leads to errors in the estimates of \( {}^G\mathbf{R}_I \) and \( {}^G\mathbf{v}_I \). Moreover, using the gyroscope measurement at an incorrect time instant introduces additional errors, and these errors accumulate over time. To correct this, the proposed method estimates the time offset \( t_d \) from the difference in radar ego-velocity induced by temporal misalignment. This difference can arise in various motion scenarios, such as rapid changes in acceleration or angular velocity. In contrast, when the platform is stationary or exhibits only mild motion, the ego-velocity change is small, making \( t_d \) difficult to be estimated. The Jacobian in Eq.~\eqref{td_jacobian} is derived based on the estimated state and IMU measurements evaluated at the shifted time \( t' \) and is used for estimating the time offset.
\section{Experiments}
\label{sec: experiments}

\subsection{Experimental Setup}
\label{sec: experimental_setup}
\begin{figure}[t]
\centering
	\includegraphics[width=\linewidth]{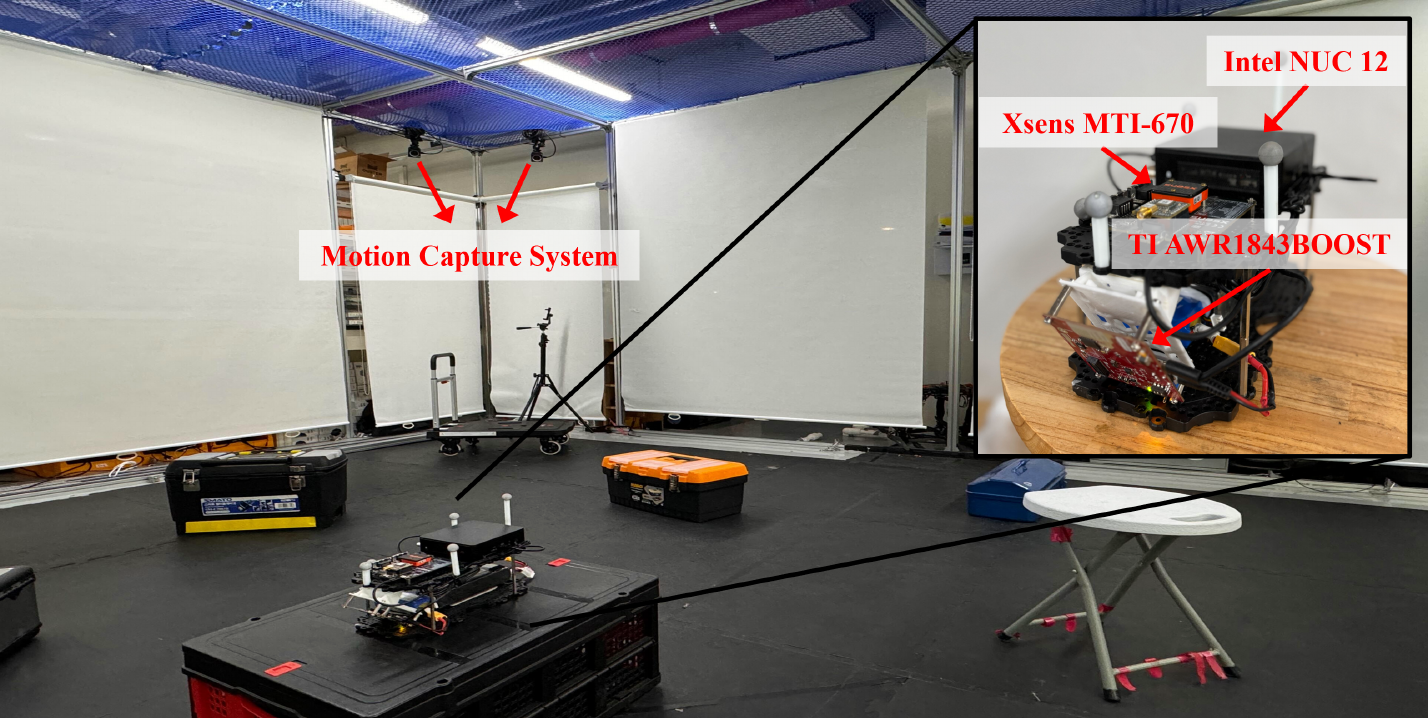}
	\vspace{-2.0em}
	\caption{The handheld platform configuration, including the radar, IMU, and onboard computer. The experiments are conducted in a room equipped with a motion capture system to obtain accurate ground truth.}
	\vspace{-0.5em}
	\label{fig2}
\end{figure}
We validate the proposed method using both simulated and real-world datasets. In the simulation, we use Gazebo with a drone equipped with a virtual IMU and a radar sensor. To simulate realistic behavior, noise is added to the IMU and radar ego-velocity measurements. A time delay of -0.15 seconds is intentionally applied to the radar timestamps to evaluate the performance of the time offset estimation.

For real-world validation, we evaluate the proposed method on three datasets, comprising a total of 15 sequences. One is our self-collected dataset, captured with a handheld platform as shown in Fig.~\ref{fig2}, while the other two are public radar datasets: ICINS2021~\cite{9470842} and ColoRadar~\cite{kramer2022coloradar}. The sensors on our platform include a Texas Instruments (TI) AWR1843BOOST radar and an Xsens MTI670 IMU. No additional hardware triggers are used between the sensors, and the sensor data is recorded using an Intel NUC i7 onboard computer. The experiments are conducted in an indoor area equipped with a motion capture system to obtain precise ground truth. The extrinsic calibration between the IMU and the radar is performed manually. To highlight the significance of temporal calibration in RIO, we design the dataset with two levels of difficulty. Sequences 1-3 involve smoother motion with smaller gyroscope changes over the time offset interval, while Sequences 4-7 exhibit greater variation, leading to a larger radar ego-velocity discrepancy and clearer impact of the time offset.
\begin{figure*}[t]
	\centering
	\includegraphics[width=\linewidth]{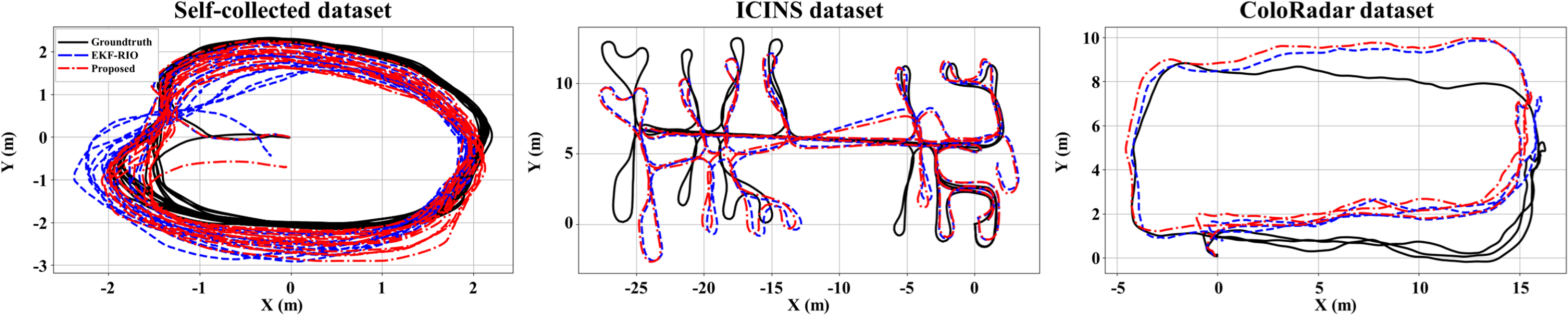}
	\vspace{-2.0em}
	\caption{Comparison of estimated trajectories with the ground truth. The \textcolor{black}{black} trajectory is the ground truth, the \textcolor{blue}{blue} one is the EKF-RIO, which does not account for temporal calibration, and the \textcolor{red}{red} one is the proposed RIO with online temporal calibration. Results are presented for Sequence 4, ICINS 1, and ColoRadar 1, representing one sequence from each of the three datasets.}
	\vspace{-0.5em}
	\label{trajectory}
\end{figure*}

In~\cite{9470842}, the ICINS2021 dataset was collected using a TI IWR6843AOP radar, an ADIS16448 IMU, and a camera. A microcontroller board was used for active hardware triggering to ensure accurate radar timing. Data was captured on both handheld and drone platforms. The handheld sequences ``carried\_1'' and ``carried\_2'' are referred to as ``ICINS 1'' and ``ICINS 2'', respectively. The drone sequences ``flight\_1'' and ``flight\_2'' are referred to as ``ICINS 3'' and ``ICINS 4'', respectively. Ground truth is provided by visual-inertial SLAM with multiple loop closures, serving as pseudo-ground truth. In~\cite{kramer2022coloradar}, the ColoRadar dataset was collected using a TI AWR1843BOOST radar, a 3DM-GX5-25 IMU, and a LiDAR on a handheld platform. No hardware synchronization was applied. The sequences ``arpg\_lab\_run0'' and ``arpg\_lab\_run1'' are referred to as ``ColoRadar 1'' and ``ColoRadar 2'', respectively, while ``ec\_hallways\_run0'' and ``ec\_hallways\_run1'' are referred to as ``ColoRadar 3'' and ``ColoRadar 4'', respectively. Ground truth is obtained via LiDAR-inertial SLAM with loop closures.
\subsection{Evaluation}
\label{sec: evaluation}
For the performance comparison, the open-source EKF-RIO \cite{9235254}, which uses the same measurement model but does not account for temporal calibration, is employed. All parameters are kept identical to ensure a fair comparison. In the proposed method, the time offset \( t_d \) is initialized to 0.0 seconds for all sequences, reflecting a typical scenario where the initial time offset is unknown. The experimental results are evaluated using the open-source tool EVO \cite{grupp2017evo}. Figure~\ref{trajectory} illustrates the estimated trajectories compared to the ground truth for visual comparison, with one representative result from each dataset. Due to the stochastic nature of the RANSAC algorithm used in radar ego-velocity estimation, the averaged results from 100 trials across all datasets are presented. We compare the root mean square error (RMSE) of both absolute pose error (APE) and relative pose error (RPE), with the RPE calculated at 10-meter intervals, both evaluated after origin alignment.
\subsubsection{Simulation}
\begin{figure}[t]
	\centering
	\includegraphics[width=\linewidth]{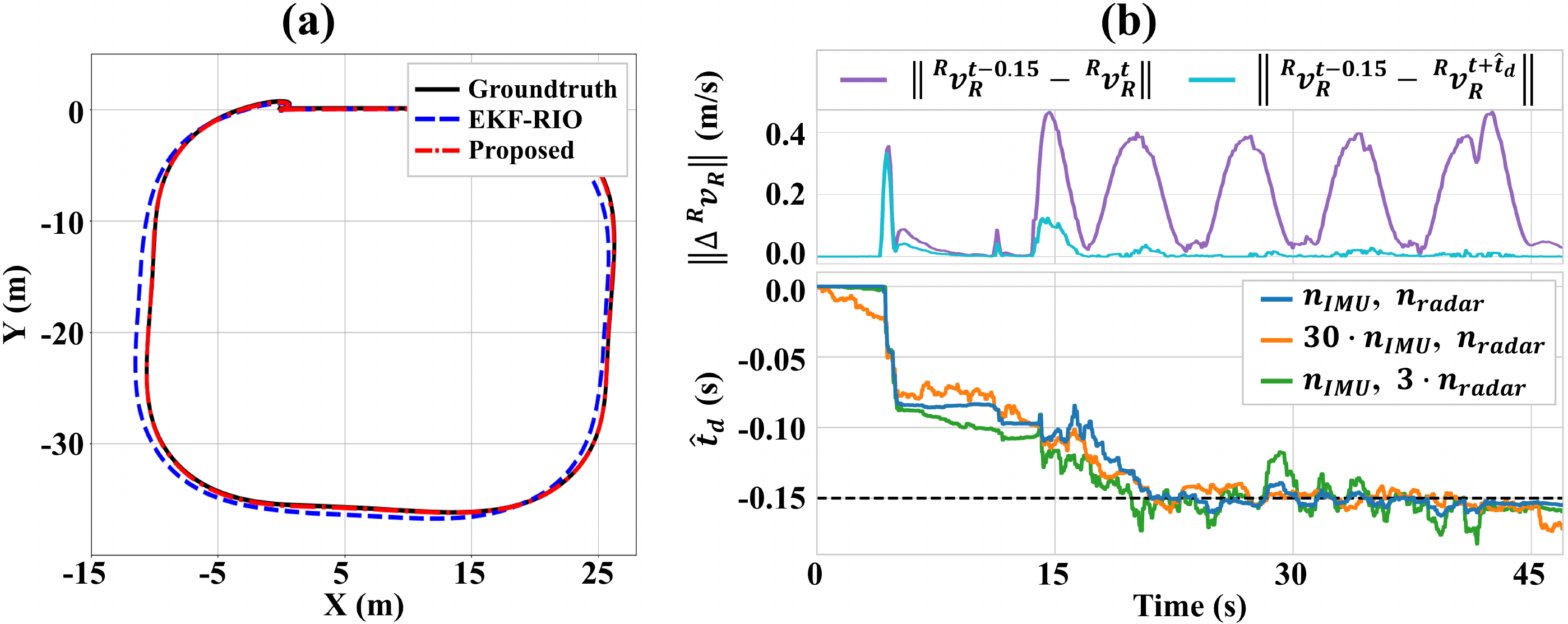}
	\vspace{-2.0em}
	\caption{Simulation results: (a) Trajectory comparison, and (b) top: difference in radar ego-velocity due to temporal shift by the time offset, and bottom: time offset estimation under different sensor noise levels.}
	\vspace{-0.5em}
	\label{sim}
\end{figure}

The simulation experiment aims to validate the performance and importance of the proposed temporal calibration under a known time offset. Figure~\ref{sim}(a) shows the trajectory comparison between EKF-RIO without temporal calibration and the proposed method, demonstrating improved accuracy by correcting temporal misalignment.

As discussed in Section~\ref{sec: online temporal calibration}, the time offset is estimated from the discrepancy in radar ego-velocity caused by temporal misalignment. The upper plot in Fig.~\ref{sim}(b) illustrates this discrepancy by comparing the radar ego-velocity at the true measurement time (i.e., $t-0.15$), the delayed timestamp (i.e., $t$), and the time adjusted by the estimated offset (i.e., $t+\hat{t}_d$). A larger discrepancy leads to a stronger correction in the estimated time offset. As the proposed method gradually compensates for the offset, this difference diminishes over time, indicating improved temporal alignment. The lower plot in Fig.~\ref{sim}(b) illustrates how increased sensor noise affects the stability of time offset estimation. Results with $30\times$ IMU noise and $3\times$ radar noise are shown for comparison. The estimation becomes more unstable with higher noise levels, especially with radar noise, since the method depends not only on IMU measurements but also on the estimated state. Accurate radar measurements are crucial, as they directly influence the state estimation in RIO.

\subsubsection{Self-Collected Dataset}
\begin{table}[t]
	\centering
	\caption{Quantitative Results of Fixed Offset and Online Estimation}
	\vspace{-0.8em}
	\label{fixed_offset}
	\resizebox{\linewidth}{!}{
		\begin{tblr}{
				cells = {c},
				cell{1}{1} = {r=2}{},
				cell{1}{2} = {r=2}{},
				cell{1}{3} = {r=2}{},
				cell{1}{4} = {c=2}{},
				cell{1}{6} = {c=2}{},
				cell{3}{1} = {r=6}{},
				cell{3}{2} = {r=5}{},
				cell{3}{5} = {fg=red},
				cell{4}{4} = {fg=red},
				cell{5}{4} = {fg=blue},
				cell{5}{5} = {fg=blue},
				cell{5}{6} = {fg=blue},
				cell{5}{7} = {fg=red},
				cell{6}{6} = {fg=red},
				cell{6}{7} = {fg=blue},
				cell{9}{1} = {r=6}{},
				cell{9}{2} = {r=5}{},
				cell{11}{4} = {fg=red},
				cell{11}{5} = {fg=blue},
				cell{11}{6} = {fg=red},
				cell{11}{7} = {fg=red},
				cell{12}{4} = {fg=blue},
				cell{12}{5} = {fg=red},
				cell{12}{6} = {fg=blue},
				cell{12}{7} = {fg=blue},
				hline{1,3,9,15} = {-}{},
				hline{2} = {4-7}{},
			}
			\textbf{Sequence} & \textbf{Method} &  \textbf{Time Offset (s)}            & \textbf{APE RMSE} &                & \textbf{RPE RMSE} &                   \\
			&                 &                                      & Trans. (m)        & Rot. (\degree) & Trans. (m)        & Rot. (\degree)    \\
			\hline
			Sequence 1        & Fixed Offset    & 0.0             & 0.985             & 1.872          & 0.264             & 1.230          \\
			&                 & -0.05           & 0.647             & 7.561          & 0.166             & 1.549          \\
			&                 & -0.10           & 0.661             & 2.438          & 0.138             & 0.948          \\
			&                 & -0.15           & 0.826             & 5.151          & \textbf{0.131}    & 1.196          \\
			&                 & -0.20           & 0.974             & 2.698          & 0.156             & 1.274          \\
			& Online Est.     & \textbf{-0.114} & \textbf{0.646}    & \textbf{0.935} & 0.132    & \textbf{0.774} \\
			Sequence 4        & Fixed Offset    & 0.0             & 1.737             & 25.885         & 0.118             & 4.074          \\
			&                 & -0.05           & 1.028             & 15.460         & 0.091             & 2.313          \\
			&                 & -0.10           & 0.635             & 4.655          & 0.061             & 0.994          \\
			&                 & -0.15           & 0.649             & 4.275          & 0.068             & 1.083          \\
			&                 & -0.20           & 0.716             & 12.461         & 0.092             & 2.526          \\
			& Online Est.     & \textbf{-0.115} & \textbf{0.610}    & \textbf{3.099} & \textbf{0.057}    & \textbf{0.944} 
		\end{tblr}
	}
	\vspace{0.3em}
	{\raggedright
		\noindent\par {\footnotesize \textsuperscript{*}The initial time offset of `Online Est.' is set to 0.0 and the converged values are shown above.}
		\noindent\par {\footnotesize \textsuperscript{**}For each sequence, the lowest error values among the fixed offsets are highlighted in \textcolor{red}{red}, and the second-lowest in \textcolor{blue}{blue}.}
		\par}
	\vspace{-1.0em}
\end{table}
The purpose of the self-collected dataset is to identify the actual time offset between the IMU and the radar and evaluate its impact on the accuracy of RIO. Since the handheld platform does not utilize a hardware trigger to synchronize the sensors, the exact time offset is unknown and must be estimated. To address this uncertainty, we evaluate the performance of fixed time offsets over a range of values to determine the interval that provides the best accuracy and estimate the likely time offset range.

\begin{figure}[t]
	\centering
	\includegraphics[width=\linewidth]{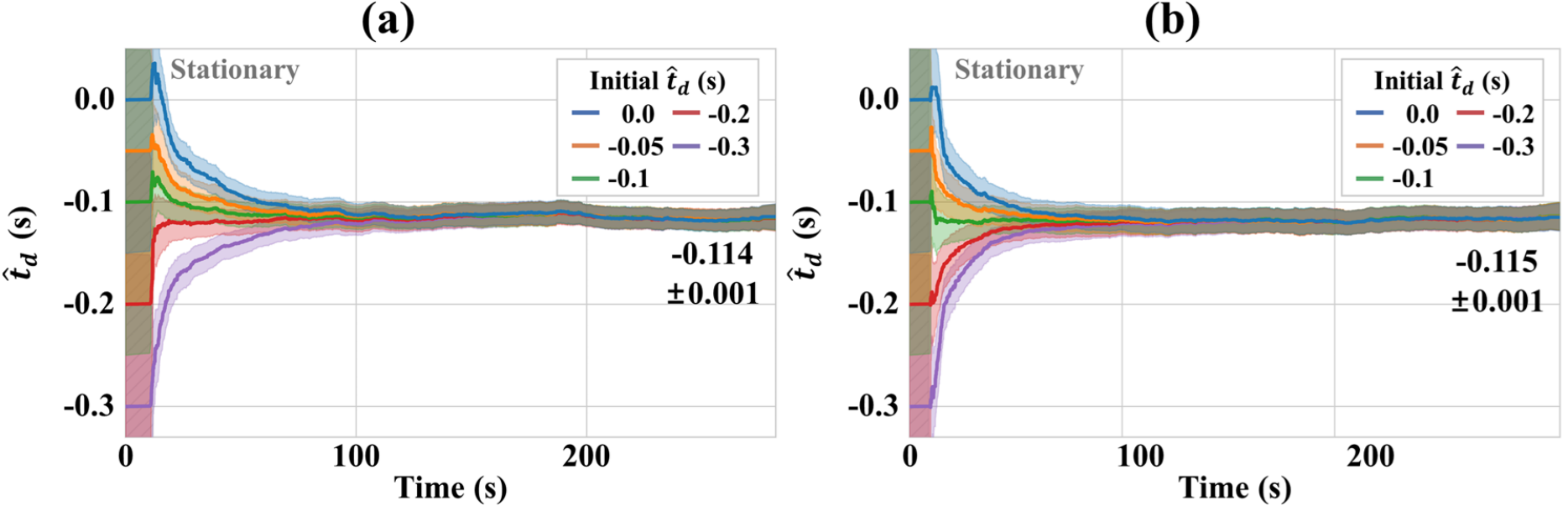}
	\vspace{-2.0em}
	\caption{Time offset estimation with 3-sigma boundaries for different initial values: (a) Sequence 1, and (b) Sequence 4.}
	\vspace{-0.5em}
	\label{sq5}
\end{figure}

As shown in Table~\ref{fixed_offset}, error values are analyzed with fixed offsets at 0.05-second intervals for Sequence 1 and Sequence 4, which feature different motion patterns. The best APE and RPE occur within the offset range of -0.10 to -0.15 seconds. The proposed method estimates the offset as -0.114 and -0.115 seconds for Sequence 1 and 4, respectively, closely matching this range. In both cases, it achieves better APE and RPE, validating the accuracy of the online estimation.
\begin{table}[t]
\centering
\caption{Quantitative Results of Comparison study on Self-collected dataset}
\vspace{-0.8em}
\label{table_self}
\resizebox{\linewidth}{!}{
\begin{tblr}{
  cells = {c},
  cell{1}{1} = {r=2}{},
  cell{1}{2} = {r=2}{},
  cell{1}{3} = {c=2}{},
  cell{1}{5} = {c=2}{},
  cell{3}{1} = {r=2}{},
  cell{5}{1} = {r=2}{},
  cell{7}{1} = {r=2}{},
  cell{9}{1} = {r=2}{},
  cell{11}{1} = {r=2}{},
  cell{13}{1} = {r=2}{},
  cell{15}{1} = {r=2}{},
  cell{17}{1} = {r=2}{},
  hline{1,3,5,7,9,11,13,15,17,19} = {-}{},
  hline{2} = {3-6}{},
}
{\textbf{Sequence }\\\textbf{(Trajectory Length, $\Delta \omega$)}} & {\textbf{Method } \textbf{($\hat{t}_d$)}} & \textbf{APE RMSE } &                & \textbf{RPE RMSE } &                \\
                                                   &                                         & Trans. (m)         & Rot. (\degree)        & Trans. (m)         & Rot. (\degree)        \\
                                                   \hline
{Sequence 1\\(177 m, 0.23 rad/s)}                              & {EKF-RIO (N/A)}                        & 0.985              & 1.872           & 0.264              & 1.230          \\
                                                   & {Ours (-0.114 s)}                      & \textbf{0.646}     & \textbf{0.935}  & \textbf{0.132}     & \textbf{0.774} \\
{Sequence 2\\(197 m, 0.14 rad/s)}                              & {EKF-RIO}                              & 2.269              & 2.161           & 0.136              & 1.414          \\
                                                   & {Ours (-0.114 s)}                      & \textbf{0.587}     & \textbf{1.650}  & \textbf{0.064}     & \textbf{0.784} \\
{Sequence 3\\(144 m, 0.12 rad/s)}                              & {EKF-RIO}                              & 1.368              & 2.331           & 0.167              & 1.347          \\
                                                   & {Ours (-0.113 s)}                      & \textbf{0.414}     & \textbf{1.140}  & \textbf{0.088}     & \textbf{0.613} \\
{Sequence 4\\(197 m, 0.32 rad/s)}                              & {EKF-RIO}                              & 1.737              & 25.885          & 0.118              & 4.074          \\
                                                   & {Ours (-0.115 s)}                      & \textbf{0.610}     & \textbf{3.099}  & \textbf{0.057}     & \textbf{0.944} \\
{Sequence 5\\(190 m, 0.29 rad/s)}                              & {EKF-RIO}                              & 2.375              & 7.702           & 0.122              & 1.600          \\
                                                   & {Ours (-0.115 s)}                      & \textbf{1.150}     & \textbf{1.304}  & \textbf{0.069}     & \textbf{0.814} \\
{Sequence 6\\(179 m, 0.28 rad/s)}                              & {EKF-RIO}                              & 1.267              & 17.907          & 0.117              & 2.828          \\
                                                   & {Ours (-0.111 s)}                      & \textbf{0.661}     & \textbf{2.551}  & \textbf{0.051}     & \textbf{0.809} \\
{Sequence 7\\(223 m, 0.26 rad/s)}                              & {EKF-RIO}                              & 2.757              & 10.092          & 0.116              & 1.863          \\
                                                   & {Ours (-0.112 s)}                      & \textbf{1.596}     & \textbf{6.039}  & \textbf{0.057}     & \textbf{1.365} \\
{Average}                                          & {EKF-RIO}                              & 1.822              & 9.707            & 0.148             & 2.051          \\
                                                   & {Ours (-0.113 s)}                      & \textbf{0.809}     & \textbf{2.388}   & \textbf{0.074}    & \textbf{0.872}   
\end{tblr}
}
\vspace{0.3em}
{\raggedright
\noindent\par {\footnotesize \textsuperscript{*}\(\Delta \omega\) is defined as the norm of the averaged gyroscope measurement over the interval \([t - 0.11, t]\), capturing the rotational motion during that period.
}
\par}
\vspace{-1.0em}
\end{table}
\begin{table}[t]
	\centering
	\caption{Comparison of Time Offset in Multi-Sensor Fusion Systems}
	\vspace{-0.8em}
	\label{time_offset_comparison}
	\begin{tabular}{|c|c|c|} 
		\hline
		\textbf{Systems} & \textbf{Sensor} & \textbf{Time Offset} \\ 
		\hline
		LiDAR-Inertial~\cite{10113826} & Velodyne VLP-32 & -0.006 s\\ 
		\hline
		Visual-Inertial~\cite{li2014online} & PointGrey Bumblebee2 & -0.047 s\\ 
		\hline
		Radar-Inertial & TI AWR1843BOOST & \textbf{-0.113 s} \\
		\hline
	\end{tabular}
	\vspace{-1.0em}
\end{table}
To evaluate the robustness of the estimation, different initial values of \( t_d \) ranging from 0.0 to -0.3 seconds are tested. Figure~\ref{sq5} illustrates the estimated time offset for each initial setting, along with the 3-sigma boundaries. As \( t_d \) is estimated from radar ego-velocity, it cannot be determined while the platform is stationary. Once the platform starts moving, the filter begins estimating \( t_d \) and quickly converges to a stable value. The filter converges to a stable time offset of -0.114 ± 0.001 seconds in Sequence 1 and -0.115 ± 0.001 seconds in Sequence 4.

Table \ref{table_self} presents the performance comparison between the proposed method with online temporal calibration and EKF-RIO across seven sequences. The proposed method outperforms EKF-RIO, significantly reducing both APE and RPE across all sequences. Specifically, it reduces APE translation error by an average of 56\%, APE rotation error by 75\%, RPE translation error by 50\%, and RPE rotation error by 57\% compared with EKF-RIO. Despite using the same measurement model, the performance improvement is achieved solely by applying propagation and updates based on a common time stream through the proposed online temporal calibration.
\begin{figure}[t]
	\centering
	\includegraphics[width=\linewidth]{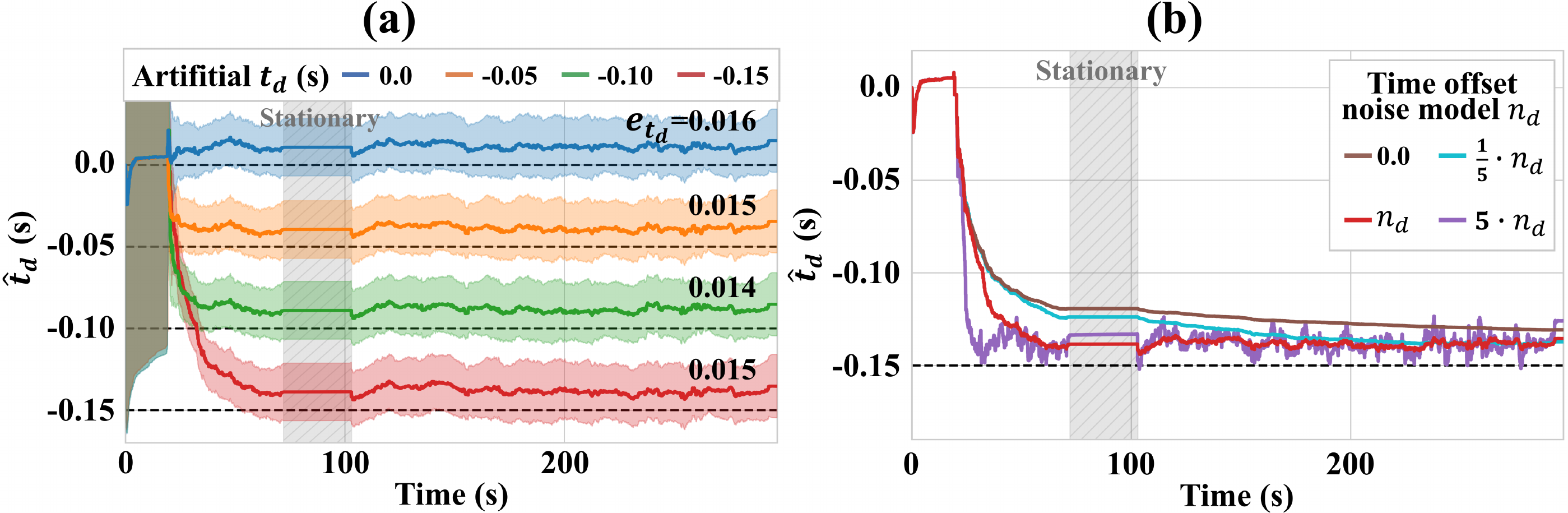}
	\vspace{-2.0em}
	\caption{Time offset estimation in ICINS 1: (a) estimation results for different artificial offsets, and (b) effect of propagation noise modeling under a delay of $-0.15$s.}
	\vspace{-0.5em}
	\label{icins1}
\end{figure}

On average, the time offset \( t_d \) is estimated to be -0.113 ± 0.002 seconds, confirming consistent temporal calibration throughout the experiments. Compared with LiDAR-inertial and visual-inertial systems, radar-inertial systems exhibit a significantly larger time offset, as shown in Table~\ref{time_offset_comparison}. Given the radar sensor rate (10 Hz), such a large time offset is significant enough to cause a misalignment spanning more than one data frame. These findings highlight the necessity of temporal calibration in RIO, which is crucial for accurate sensor fusion and reliable pose estimation in real-world applications.

\subsubsection{Open Datasets}
The ICINS dataset provides radar data synchronized via a hardware trigger, allowing it to serve as a reference for validating time offset estimation. To evaluate the proposed method, we apply artificial delays to the radar timestamps. As shown in Fig.~\ref{icins1}(a), the proposed method estimates the time offsets with an average error, \(e_{t_d}=|t_d^{\text{true}}-\hat{t}_d|\), of approximately 0.015 seconds. Figure~\ref{icins1}(b) shows the effect of the propagation noise on time offset estimation. Increasing the noise level accelerates convergence but reduces stability, while a smaller noise improves stability at the cost of slower convergence. These results highlight the necessity of noise modeling and emphasize the importance of appropriately tuning the noise parameter to balance responsiveness and robustness in time offset estimation. The results for the ICINS dataset without artificial delay are presented in Table~\ref{opendataset}. The proposed method estimates the time offset as 0.016 ± 0.003 seconds. Despite the slight deviation from the expected value of 0.0 seconds, the method achieves comparable or improved APE and RPE performance compared to EKF-RIO. Although the ICINS dataset includes hardware-triggered signals for the radar, it does not provide such triggering for the IMU. As defined in Eq.~\eqref{time_offset}, the estimated positive time offset is attributed to this IMU delay, explaining the difference from the expected value.

The ColoRadar dataset, widely used for performance comparison in the RIO field, is utilized to assess if the proposed method generalizes well across different datasets. As shown in Table~\ref{opendataset}, the proposed method also demonstrates performance improvements over EKF-RIO in terms of both APE and RPE on average. However, the extent of improvement is smaller compared with the self-collected dataset, which can be explained by differences in motion characteristics. This is attributed to the lower rotational motion present in the ColoRadar dataset, as indicated by smaller values of \( \Delta \omega \), resulting in reduced impact of the time offset. Nonetheless, the proposed method achieves 33\% reduction in RPE translation error, demonstrating its effectiveness even in this less challenging trajectory. On average, the time offset \( t_d \) is estimated to be -0.111 ± 0.003 seconds, similar to the time offset found in the self-collected dataset. This consistency is likely due to the use of the same radar sensor model in both datasets, further validating the reliability of the proposed method across different environments.

\begin{table}[t]
\centering
\caption{Quantitative Results of Comparison study on Open datasets}
\vspace{-0.8em}
\label{opendataset}
\resizebox{\linewidth}{!}{
\begin{tblr}{
  cells = {c},
  cell{1}{1} = {r=2}{},
  cell{1}{2} = {r=2}{},
  cell{1}{3} = {c=2}{},
  cell{1}{5} = {c=2}{},
  cell{3}{1} = {r=2}{},
  cell{5}{1} = {r=2}{},
  cell{7}{1} = {r=2}{},
  cell{9}{1} = {r=2}{},
  cell{11}{1} = {r=2}{},
  cell{13}{1} = {r=2}{},
  cell{15}{1} = {r=2}{},
  cell{17}{1} = {r=2}{},
  cell{19}{1} = {r=2}{},
  cell{21}{1} = {r=2}{},
  hline{1,3,5,7,9,11,13,15,17,19,21,23} = {-}{},
  hline{2-3} = {3-6}{},
}
{\textbf{Sequence }\\\textbf{(Trajectory Length, $\Delta \omega$)}}       & \textbf{Method ($\hat{t}_d$)} & \textbf{APE RMSE}        &\textbf{RPE RMSE}  &                           \\
                        &                               & Trans. (m)               & Rot. (\degree)                            & Trans. (m)              & Rot. (\degree)            \\
                        \hline
{ICINS 1\\(295 m)}      & EKF-RIO (N/A)                 & 1.959                    & 10.694                                    & \textbf{0.093}          & \textbf{0.896}            \\
                        & Ours (0.016 s)                & \textbf{1.922}           & \textbf{10.135}                           & 0.098                   & 0.918                     \\
{ICINS 2\\(468 m)}      & EKF-RIO                       & 3.830                    & 23.151                                    & \textbf{0.114}          & 1.289                     \\
                        & Ours (0.013 s)                & \textbf{3.198}           & \textbf{19.235}                           & 0.121                   & \textbf{1.076}            \\
{ICINS 3\\(150 m)}      & EKF-RIO                       & \textbf{1.502}           & \textbf{9.905}                            & 0.130                   & \textbf{1.512}            \\
                        & Ours (0.015 s)                & 1.530                    & 10.189                                    & \textbf{0.126}          & 1.553                     \\
{ICINS 4\\(50 m)}       & EKF-RIO                       & \textbf{0.213}           & \textbf{2.091}                            & \textbf{0.076}          & \textbf{0.923}            \\
                        & Ours (0.019 s)                & 0.216                    & 2.098                                     & 0.081                   & \textbf{0.923}            \\
Average                 & EKF-RIO                       & 1.876                    & 11.460                                    & \textbf{0.103}          & 1.155                     \\
                        & Ours (0.016 s)                & \textbf{1.716}           & \textbf{10.414}                           & 0.106                   & \textbf{1.117}            \\
                        \hline
{ColoRadar 1\\(178 m, 0.02 rad/s)} & EKF-RIO (N/A)                 & 6.556                    & \textbf{1.354}                   & 0.182                   & \textbf{1.071}          \\
                        		   & Ours (-0.110 s)               & \textbf{6.173}           & 1.382                            & \textbf{0.155}          & 1.188                   \\
{ColoRadar 2\\(197 m, 0.08 rad/s)} & EKF-RIO                       & \textbf{4.747}           & 1.238                            & 0.372                   & 1.375                   \\
                                   & Ours (-0.114 s)               & 4.826                    & \textbf{0.960}                   & \textbf{0.292}          & \textbf{1.180}          \\
{ColoRadar 3\\(197 m, 0.07 rad/s)} & EKF-RIO                       & \textbf{8.307}           & 1.969                            & 0.259                   & 1.015                   \\
                                   & Ours (-0.108 s)               & 8.550                    & \textbf{1.852}                   & \textbf{0.221}          & \textbf{0.879}          \\
{ColoRadar 4\\(144 m, 0.04 rad/s)} & EKF-RIO                       & 12.111                   & 2.815                            & 0.488                   & 1.263                   \\
                                   & Ours (-0.112 s)               & \textbf{11.946}          & \textbf{2.756}                   & \textbf{0.200}          & \textbf{1.116}          \\
Average                            & EKF-RIO                       & 7.930                    & 1.844                            & 0.325                   & 1.181                   \\
                                   & Ours (-0.111 s)               & \textbf{7.874}           & \textbf{1.737}                   & \textbf{0.217}          & \textbf{1.091}          
\end{tblr}
}
\vspace{0.3em}
{\raggedright
	\noindent\par {\footnotesize \textsuperscript{*}\(\Delta \omega\) is omitted for the ICINS sequences, as they are reported to be time-synchronized.
	}
	\par}
\vspace{-1.0em}
\end{table}

\section{Conclusions and Future Work}
\label{sec: conclusion}
In this paper, we proposed an EKF-based RIO framework with online temporal calibration. To ensure accurate sensor time synchronization during IMU and radar sensor fusion, the time offset between sensors is estimated from radar ego-velocity, which is derived from a single radar scan. This approach avoids the potential risks of finding correspondences between consecutive radar scans and, being independent of radar point cloud density, offers flexibility for use with various types of radar sensors. By leveraging temporal calibration, sensor measurements are aligned to a common time stream. This allows propagation and measurement updates to be applied at the correct time, improving overall performance. Extensive experiments across multiple datasets demonstrate the effectiveness of time offset estimation and provide a detailed analysis of its impact on overall performance.

Several challenges remain in multi-sensor fusion state estimation using radar systems. One issue is the reliance on manually calibrated sensor extrinsic parameters in many studies, which can lead to inaccuracies. We will focus on spatiotemporal calibration between sensors to further improve the accuracy and robustness of multi-sensor fusion systems.

\addtolength{\textheight}{-12cm}   




\end{document}